\documentclass[10pt, a4paper]{article}

\usepackage{lrec-coling2024} 

\usepackage{graphicx}
\usepackage{booktabs}
\usepackage{makecell}
\usepackage{hyperref}

\title{Interpreting Themes from Educational Stories}

\name{Yigeng Zhang\textsuperscript{1}, Fabio A. González\textsuperscript{2}, Thamar Solorio\textsuperscript{1,3}} 

\address{\textsuperscript{1}University of Houston, Houston, USA\\\textsuperscript{2}Universidad Nacional de Colombia, Bogotá, Colombia\\\textsuperscript{3}MBZUAI, Masdar City, United Arab Emirates \\
         \textsuperscript{1}\texttt{\{yzhang168,tsolorio\}@uh.edu}, 
         \textsuperscript{2}\texttt{fagonzalezo@unal.edu.co}}

\abstract{
Reading comprehension continues to be a crucial research focus in the NLP community. Recent advances in Machine Reading Comprehension (MRC) have mostly centered on literal comprehension, referring to the surface-level understanding of content. In this work, we focus on the next level - interpretive comprehension, with a particular emphasis on inferring the themes of a narrative text. We introduce the first dataset specifically designed for interpretive comprehension of educational narratives, providing corresponding well-edited theme texts. The dataset spans a variety of genres and cultural origins and includes human-annotated theme keywords with varying levels of granularity. We further formulate NLP tasks under different abstractions of interpretive comprehension toward the main idea of a story. After conducting extensive experiments with state-of-the-art methods, we found the task to be both challenging and significant for NLP research. The dataset and source code have been made publicly available to the research community at \href{https://github.com/RiTUAL-UH/EduStory}{\texttt{https://github.com/RiTUAL-UH/EduStory}}.
 \\ \newline \Keywords{Corpus, Document Classification, Text Categorization, Question Answering} }

\begin{document}

\maketitleabstract

\section{Introduction}

Reading and understanding are fundamental aspects of human intellectual activity. Reading comprehension is one of the many abilities that AI is anticipated to develop on par with humans. The NLP community has dedicated substantial efforts to machine reading comprehension (MRC) research, resulting in significant advancements in models' reading capabilities. 
From an educational research standpoint, reading comprehension is divided into three levels: literal comprehension, inferential/interpretive comprehension, and critical/evaluative comprehension (\cite{herber1978}, further developed by \cite{vacca1998}). The first level involves understanding direct and explicit information extracted from a text, including facts, vocabulary, events, and other stated information. The second level demands that readers make inferences from contextual information, such as deducing cause and effect or determining the main idea. 
The third level transcends the text, requiring readers to integrate their own opinions and critically analyze the content or assess a viewpoint.

\begin{figure}[t]
\centering
  \includegraphics[width=0.99\linewidth]{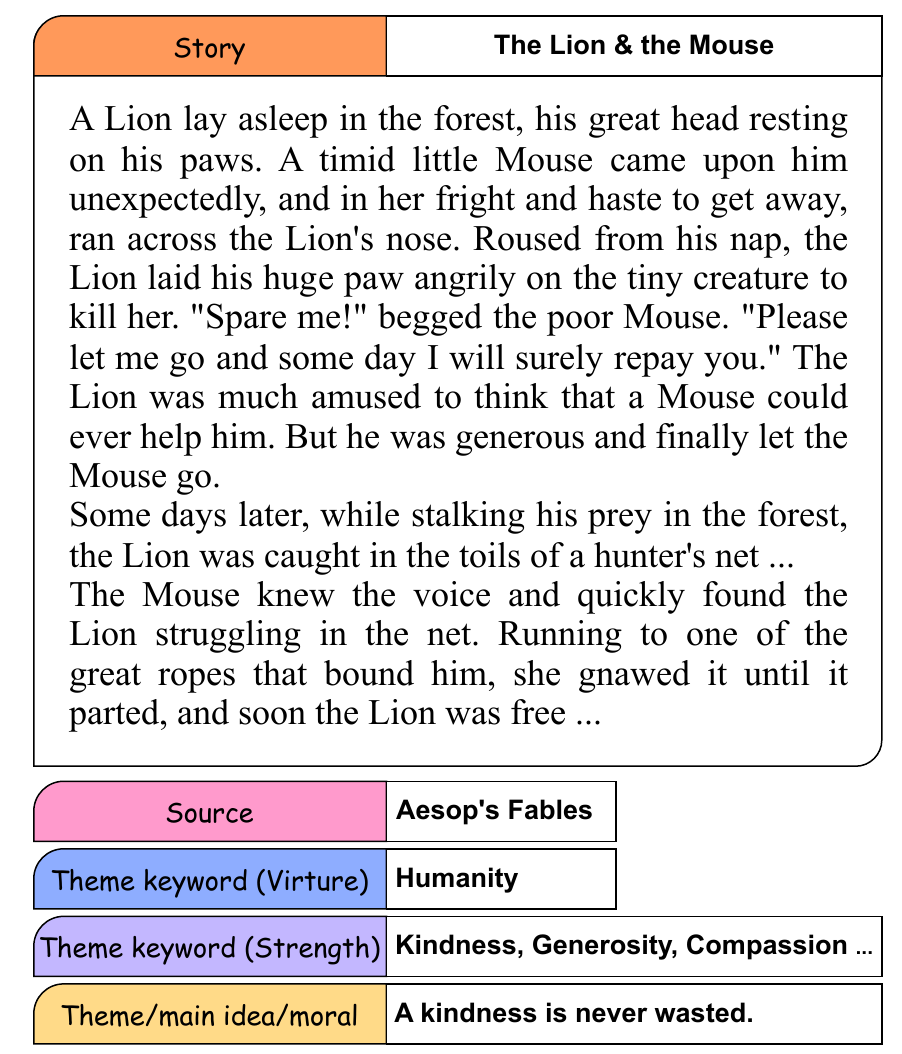}
  \caption{An example of theme interpretation.}
  \label{fig:storyexample}
\end{figure}

Current NLP research does not explicitly regard reading comprehension from different levels or distinguish between them, with the majority of MRC research focusing on the literal level \cite{richardson-etal-2013-mctest, kocisky-etal-2018-narrativeqa, saha-etal-2018-duorc}. However, in real-world learning environments, mere word decoding and literal matching are inadequate. Recognizing the inherent meaning of a text or its implied information remains an area of ongoing study. This work concentrates on a novel research problem: interpreting themes from text using NLP methods. This topic falls within the second level, interpretive comprehension. 
A theme goes beyond a simple summary of the story's plot or character actions. Instead, it reflects deeper insights and conveys the key message that is implied within the context. This complexity requires that NLP models not only process the context but also make inferences and interpret the theme or main idea, which is often not explicitly stated in the text.

To further explore and gain empirical knowledge on this research problem, we choose educational stories as our context. These narrative texts, such as fables and folktales, often convey a lesson via a series of events with a clear consequence. These stories are widely embraced by individuals from relatively diverse cultural backgrounds and knowledge levels and are commonly used as children's bedtime reading.
For each story, the theme sentence(s) (main idea/lesson/moral/meaning) is often provided by the story's author or editor. We use our best efforts to gather educational stories and create a dataset of high-quality English story-theme pairs from various sources and cultural backgrounds. Figure \ref{fig:storyexample} depicts an exemplar story with its attributes.

Existing language resources for narrative comprehension, such as those presented in \cite{xu-etal-2022-fantastic} and \cite{Zhao2023storyqa}, have been designed primarily for explicit and implicit question answering and they do not focus on the comprehension of story themes. Additionally, these datasets tend to be not only limited in size but also lack diversity in their sources. To address this gap, we probe further into the story content and characterize the challenge of theme interpretation across various NLP abstractions. Given the challenge of interpretive comprehension, we outline tasks according to their levels of difficulty. First, we propose to investigate \textbf{theme identification}. Educational story themes are categorized based on values from positive psychology, character strengths, and virtues. The task is formulated as story classification at the theme keyword level, such as wisdom and integrity. Next, we examine \textbf{story-theme matching}, where a story is given, and its theme sentence must be found within a collection of theme sentences, or vice versa. This task involves story-theme or theme-story retrieval. Additionally, we investigate \textbf{story reading comprehension on themes}. Similar to typical MRC or Q\&A tasks, we design multiple-choice problems on themes given a story. Finally, we conduct exploratory research on \textbf{theme generation}. By leveraging recent advances in pretrained large language models (LLMs), we explore the capability of generating accurate theme text from a given story.

\begin{table*}[h!]
\centering
\resizebox{0.95\textwidth}{!}{%
\begin{tabular}{cccccccc}
\Xhline{3\arrayrulewidth}
\multicolumn{5}{c|}{\textbf{Wisdom and Knowledge}  228}                                                                                                                             & \multicolumn{3}{c}{\textbf{Humanity}  59}                                                        \\ \hline
\multicolumn{1}{c|}{Creativity}   & \multicolumn{1}{c|}{Curiosity}   & \multicolumn{1}{c|}{Judgment}  & \multicolumn{1}{c|}{Learning} & \multicolumn{1}{c|}{Perspective}  & \multicolumn{1}{c|}{Love}        & \multicolumn{1}{c|}{Kindness} & Social Intelligence \\ 
\multicolumn{1}{c|}{3}            & \multicolumn{1}{c|}{0}           & \multicolumn{1}{c|}{26}        & \multicolumn{1}{c|}{0}        & \multicolumn{1}{c|}{199}          & \multicolumn{1}{c|}{16}          & \multicolumn{1}{c|}{31}       & 12                  \\ \Xhline{3\arrayrulewidth}
\multicolumn{5}{c|}{\textbf{Transcendence}  25}                                                                                                                                     & \multicolumn{3}{c}{\textbf{Justice}  22}                                                         \\ \hline
\multicolumn{1}{c|}{Appreciation} & \multicolumn{1}{c|}{Gratitude}   & \multicolumn{1}{c|}{Hope}      & \multicolumn{1}{c|}{Humor}    & \multicolumn{1}{c|}{Spirituality} & \multicolumn{1}{c|}{Citizenship} & \multicolumn{1}{c|}{Fairness} & Leadership          \\ 
\multicolumn{1}{c|}{1}            & \multicolumn{1}{c|}{20}           & \multicolumn{1}{c|}{2}         & \multicolumn{1}{c|}{0}        & \multicolumn{1}{c|}{2}            & \multicolumn{1}{c|}{17}          & \multicolumn{1}{c|}{4}        & 1                   \\ \Xhline{3\arrayrulewidth}
\multicolumn{4}{c|}{\textbf{Courage}  57}                                                                                                       & \multicolumn{4}{c}{\textbf{Temperance}  60}                                                                                          \\ \hline
\multicolumn{1}{c|}{Bravery}      & \multicolumn{1}{c|}{Persistence} & \multicolumn{1}{c|}{Integrity} & \multicolumn{1}{c|}{Vitality} & \multicolumn{1}{c|}{Forgiveness}  & \multicolumn{1}{c|}{Humility}    & \multicolumn{1}{c|}{Prudence} & Self-Regulation     \\ 
\multicolumn{1}{c|}{5}            & \multicolumn{1}{c|}{23}          & \multicolumn{1}{c|}{29}        & \multicolumn{1}{c|}{0}        & \multicolumn{1}{c|}{2}            & \multicolumn{1}{c|}{25}          & \multicolumn{1}{c|}{17}       & 16                  \\ \Xhline{3\arrayrulewidth}
\end{tabular}%
}
\caption{Distribution of character strengths and virtues across themes. \emph{Learning} represents \emph{Love of Learning}, \emph{Forgiveness} represents \emph{Forgiveness and Mercy}, \emph{Humility} represents \emph{Humility and Modesty}, and \emph{Appreciation} represents \emph{Appreciation of Beauty and Excellence}.}
\label{tab:strength distribution}
\end{table*}

To assess how challenging the proposed theme interpreting tasks are, we designed and conducted experiments using different machine learning (ML)-based methods, covering both conventional ML models and large language model (LLM)-based techniques. Experimental results on the classification, text retrieval, and MRC tasks show that interpreting themes from narrative text is still challenging even with state-of-the-art LLM-based methods. We further use human judges to evaluate the LLM-generated theme sentences. The evaluation shows the strong capability of state-of-the-art LLMs to a certain extent, however, LLMs are far from perfect at interpreting a story theme that human judges can easily understand.

In sum,  the contribution of this work is summarized as follows:
\begin{itemize}
    \item Our work serves as an initial call to the community, urging further exploration and reflection on MRC issues from different levels. Specifically, we introduce the concept of \emph{theme interpretation} as a task in NLP, framed within the context of inferential/interpretive reading comprehension.
    \item We formulate the task comprehensively from various research aspects of NLP and provide extensive empirical research and analysis.
    \item We publish the first dataset in theme interpreting for the community\footnote{\href{https://github.com/RiTUAL-UH/EduStory}{\texttt{https://github.com/RiTUAL-UH/EduStory}}.}, which offers rich value for further investigation and development.
\end{itemize}

\section{\emph{EduStory}: the dataset}
To highlight the importance of theme interpretation and establish a benchmark, we introduce \emph{EduStory}, the first dataset specifically created for interpretive/inferential comprehension of themes in narrative text. We use educational stories from different eras and cultural backgrounds as the context and their corresponding themes. In this work, we surveyed various types of stories and went through multiple stages of data collection and annotation.

\subsection{Educational stories}
Educational stories utilized in creating the dataset are those that employ narratives to illustrate a point or teach a lesson to the reader. These narratives are typically written in plain language and clearly depict the characters' actions. Lessons are often conveyed to readers through positive or negative outcomes corresponding to character movements, further presenting educational main ideas, such as the importance of being kind to others and the harm of dishonesty.
The main ideas of the stories are hardly stated directly in the context. Readers must look beyond the literal words and employ reasoning skills to comprehend and extract insights from the narrative using their knowledge and common sense. 
This story collection covers a wide range of literary genres, which are not limited to the following: 
\begin{itemize}
    \item \textbf{Fables: } Fables are tales that mainly employ anthropomorphic animals as characters, placed in fantastical scenarios that teach ethical lessons. The purpose of this genre is to offer moral guidance through captivating narratives. 
    \item \textbf{Folk Stories: } Folk stories often originate in a particular culture or region. They are spread among people over generations as a unique medium of education. Folktales, legends, fairy tales, and more usually belong to this category and they typically have an educational message. 
    \item \textbf{Idiom Stories: } Idiom stories illuminate the origins or interpretations of idioms from a specific language or culture. These narratives often illustrate a memorable event to convey a moral lesson, thereby demonstrating the integral connection between language and morality. 
    \item \textbf{Miscellaneous: } This category comprises stories that weave educational narratives, although they may not conform to a specific genre. It includes ones that may be called moral tales, success stories, and inspiring stories.
\end{itemize}
 These stories may portray scenes of enlightenment or individuals who have made notable achievements. The educational theme is often delivered by a key character in a specific scene or the narratives emphasizing the admirable qualities of successful individuals, thereby inspiring readers towards virtues such as hard work and open-mindedness.


\begin{table*}[h!]
\centering
\resizebox{0.99\textwidth}{!}{%
\begin{tabular}{llllll}
\hline
Dataset     & Context type  & Number of articles & Task                 & Answer     & Level of RC  \\ \hline
MCTest \cite{richardson-etal-2013-mctest}      & Narrative     & 500                & Fact check           & In context & Literal      \\
SQuAD \cite{rajpurkar-etal-2016-squad}       & Informational & 536                & Fact check           & In context & Literal      \\
NarrativeQA \cite{kocisky-etal-2018-narrativeqa} & Narrative     & 1572               & Fact check           & In context & Literal      \\
DuoRC \cite{saha-etal-2018-duorc}       & Narrative     & 7680               & Fact check           & In context & Literal      \\
FairytaleQA  \cite{xu-etal-2022-fantastic}       & Narrative     & 278               & Mixed QA         & In context/implicit & Literal/interpretive     \\
StoryQA \cite{Zhao2023storyqa}       & Narrative     & 148               & Mixed QA          & In context/implicit & Literal/interpretive      \\
EduStory (this work)   & Narrative     & 580/451                & Theme interpretation & Implicit   & Interpretive \\ \hline
\end{tabular}%
}
\caption{Comparison across datasets with different contexts and levels of reading comprehension.}
\label{tab:dataset compare}
\end{table*}

\subsection{Data collection}
We use our best efforts to collect story-theme pairs with free access to the Internet. The educational story search includes but is not limited to, fables, moral stories, folk stories (folktales/legends), children's stories, and inspiring stories. All stories are written in plain language and designed to deliver educational value. More importantly, each story is accompanied by a piece of text that reveals the theme or main idea. 
While we limited our search to stories written in English, we tried to collect stories from various cultural backgrounds and different ages aiming for a relatively diverse representation. Our efforts resulted in a collection of 580 story-theme pairs, with their sources recorded. We further manually filtered out the stories with overlapping storylines and main ideas but with rather different narratives, and this resulted in 451 unique story-theme pairs. 
Nevertheless, we retained the 129 pairs with duplicates, recognizing their value as diverse, human-crafted language resources. 

\subsection{Theme keywords and annotation}
Since the stories are composed of educational values and aim to teach people, we propose organizing the stories by specific human values conveyed in the themes. For example, when parents tell the story of \emph{the shepherd and the wolf}, they expect their child to learn the importance of honesty and develop a sense of integrity. Educators, such as parents and teachers, hope that children will learn and build good character traits from stories, which require them to interpret the main idea of the story. We, therefore, introduce the taxonomy of favorable traits from positive psychology. In foundational work by Peterson and Seligman \cite{peterson2004character}, they loosely categorize character virtues into six categories with specific character strengths. We designed an annotation plan based on this principled taxonomy.

We use a hierarchical annotation and auditing scheme to label the theme keywords for the stories. 
First, two annotators are asked to read and understand the six character virtues and 24 strengths in the book. Afterwards, they work individually to read the stories and assign virtue and strength labels, basing their decisions on their best interpretation of the theme from the narrative and the original theme sentences.
The first round of annotation results in a Cohen's Kappa score of $\kappa = 0.30$. As expected, this indicates a lower level of annotation agreement, reflecting the subjective nature of human theme interpretation in educational stories.
We further introduce an iterative auditing process to gain additional views on theme interpretation. 
The first auditor independently reviews every story where the two annotators disagree on the theme keywords.
The auditor then determines the most fitting label for the story. This 'auditor-decided' label is then used as the gold label for the story theme.
If neither of the annotations satisfies the auditor, the auditor will provide a third opinion on the story theme. Next, a second auditor gets involved and focuses on disagreements between the two annotators and the first auditor. The second auditor either makes their best judgment among the three keywords or offers a fourth opinion. If there is still any disagreement, we go through this iterative process by introducing new auditors to determine the gold label. In our practice, two auditors can successfully resolve disagreements. It is important to note that these resulting annotations should not be viewed as the 'gold standard' for story interpretation. We will release annotations from all annotators and auditors to showcase the diversity of interpretations, which can serve as useful indicators for further studies. More details about the human
annotators, auditors, and judges can be found in Appendix section \ref{appendix:annotation}.

\subsection{Data analysis}
The distribution of themes is uneven based on the annotation. More than half of the themes belong to \emph{Wisdom} because many of these stories teach readers a specific life lesson rather than a definitive virtue or strength, such as \emph{integrity} and \emph{humility}. A detailed category distribution is shown in Table \ref{tab:strength distribution}.

\begin{figure}[h]
\centering
  \includegraphics[width=0.99\linewidth]{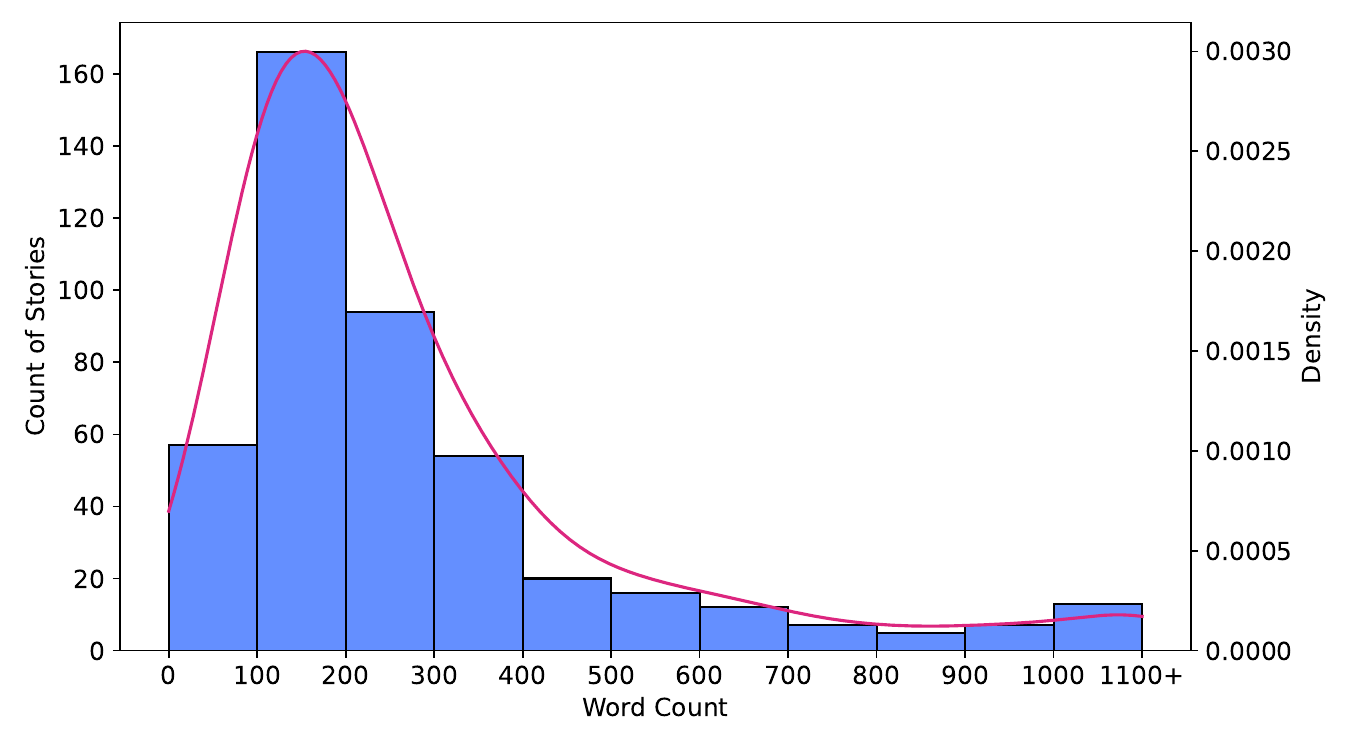}
  \caption{The distribution of word count of the story collection.}
  \label{fig:length}
\end{figure}

The stories are mostly short in length. The average length of all stories is 284 words while the median length is 201. A majority ($82\%$) of the stories have less than 400 words. A detailed text length distribution is shown in Figure \ref{fig:length}.

\begin{figure}[h]
\centering
  \includegraphics[width=0.99\linewidth]{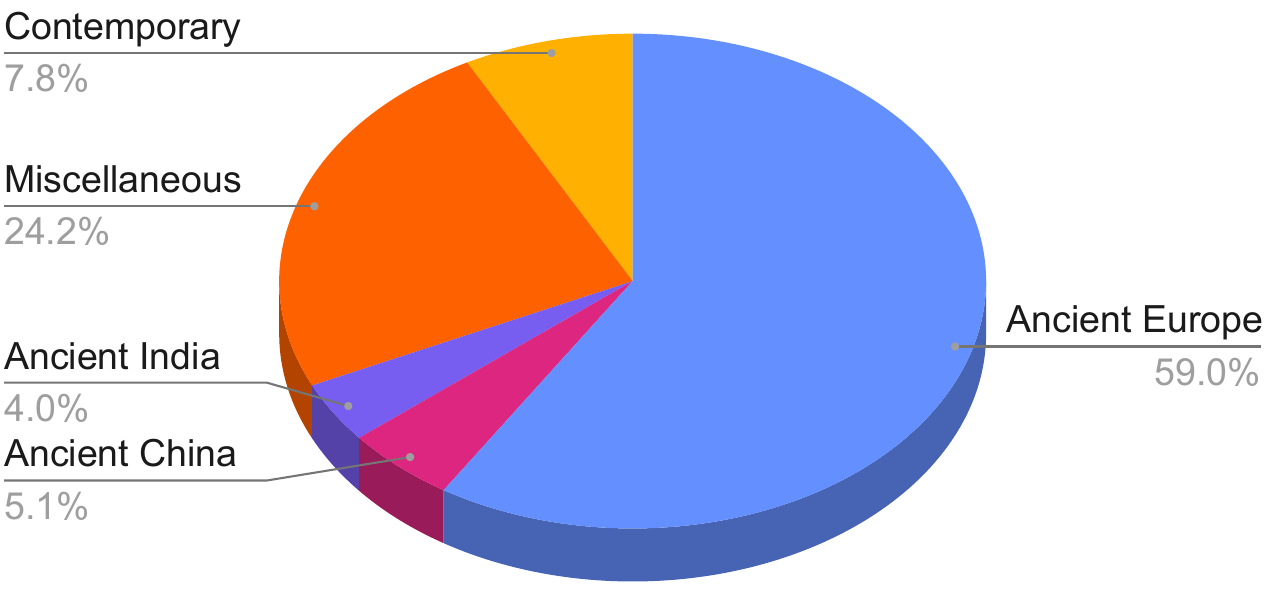}
  \caption{The statistical plot of the cultural origins of the stories in the dataset.}
  \label{fig:origin}
\end{figure}

One of the standout features of this dataset is the diverse cultural origins of the stories. Firstly, we sourced and curated stories from various versions of \emph{Aesop's Fables} found online, which resulted in over half of our collection tracing back to ancient Europe. It is important to acknowledge that not every fable associated with Aesop may be his original work. Many stories attributed to him may have unclear authorship. Given the missing evidence and the challenges of individual verification, we tentatively accept any source that labels a story as one of \emph{Aesop's Fables}. Therefore, we use the term \emph{Ancient Europe} as a loose categorization to denote the source of these stories. In addition, we have managed to gather educational stories from ancient China and India, including fables, folk tales, and idiom stories. We have also made our best effort to source other educational narratives from the open internet, including children's stories, contemporary inspirational stories, and success stories of celebrated individuals who have achieved significant accomplishments. Figure \ref{fig:origin} shows a detailed proportional representation of stories' cultural origins.

\subsection{Comparing to relevant MRC datasets}
MRC and QA have been important and popular research topics in NLP and there are many existing language resources. Existing narrative MRC datasets focus on finding specific facts and inferences in one story or plot. Typical questions such as \emph{``What did James do after he ordered the fries? ''}, \emph{``Why was the Boy so greedy?''} consists of the majority of the reading comprehension problems in the dataset. In table \ref{tab:dataset compare}, we compare \emph{EduStory} to several relevant datasets to give comprehensive information on the positioning of this work.

The primary objective of \emph{EduStory} is to evaluate the NLP capability of interpreting themes in narrative text. Although the question is as simple as \emph{``What is the main idea of the story?''}, the answer can be hardly found in the context unless one integrates information across the narrative and uses intrinsic knowledge to make inferences. This task strictly lies at the second level of reading comprehension, which means it requires more comprehensive reasoning abilities than fact-checking in context. 

\subsection{Value of further development}
The \emph{EduStory} dataset holds diverse possibilities for further development. From an NLP research perspective, additional annotations can be applied to design other MRC questions, including but not limited to literal matching and other inferential understanding. Simultaneously, \emph{EduStory} also provides resources for story generation studies based on educational themes, both from keywords and theme sentences. For educational research, this dataset can serve as a benchmark for student comprehension evaluation. Educators can leverage AI methods to compose new stories or new questions for teaching purposes. \emph{EduStory} also contributes to the fields of positive psychology and moral education. Our annotation serves as a useful reference for educators to select suitable teaching contexts. For instance, educational stories in this dataset are relatively diverse, and discovering encouragement and punishment plots and consequences is a direction for developing constructive teaching strategies and storytelling AI applications.

\section{Theme keyword identification}
The task of theme keyword identification is to assign a predefined theme keyword from a collection to a piece of the story. We introduce theme keywords of educational values from positive psychology. We have annotations on 6 higher-level classes of character virtues and 24 fine-grained character strengths as discussed in Section 2. The task is formulated as a typical multi-class text classification problem.
More formally, in a collection of $N$ labeled story-theme keyword pairs:

$ S =\{(x_1, y_1), (x_2, y_2), ..., (x_n, y_n)\}$, 

where $x_{i}$ is one piece of the story and $y_{i}$ is a theme keyword from a set of $K$ classes. The learning objective is either to build up adequate decision boundaries of different classes or to produce the desirable answer from model generation.
\subsection{Experiments}
The goal of the theme keyword identification experiment is to evaluate the performance of various supervised learning-based text classification models using different textual feature representations. We use the macro F1 score as the classification performance metric. We apply various supervised learning-based text classification models using different textual feature representations:

\textbf{TF-IDF:} For the sparse vector representation method, we compute TF-IDF vectors to represent each story passage and then apply a linear SVM to perform classification.

\textbf{Bag-of-word-vectors:} For dense vector representation, distributed word vector representations are used to vectorize each word. We take the average vector as a story passage representation and use SVM as the classifier. Here we use GloVe \cite{pennington-etal-2014-glove} as word vector representations.

\textbf{TextCNN:} We use a sequence of dense vectors to represent a passage and a convolutional neural network (CNN) to extract features and a linear layer to perform classification. Here we use the TextCNN \cite{kim-2014-convolutional} model.

\textbf{BERT: } Pretrained Transformer-based language models have been proven to be effective in various NLP tasks. We choose BERT \cite{devlin-etal-2019-bert} as one of the experiment models.

\textbf{Prompt tuning using LMs: } Prompting has shown great potential in recent research. We apply the instruction-finetuned T5 model, Flan-T5 \cite{chung2022scaling}, to perform classification with task-specific prompt tuning. We manually designed a classification template, in which we list all the keywords as textual prompting and add a prompt sentence that is the best theme keyword to describe the story. Then finetune the model to conditionally generate the correct theme keyword.

The experiments of classification on character virtues and strengths are carried out separately. Table \ref{tab:theme keyword} presents the classification performance in F1 scores for each method. This work's experiment and implementation details can be found in Appendix section \ref{appendix:experiment}.

\begin{table}[]
\centering
\resizebox{0.72\columnwidth}{!}{%
\begin{tabular}{lcccc}
\toprule[1.2pt]
        & \multicolumn{2}{c}{Virtue} & \multicolumn{2}{c}{Strength} \\ \cmidrule(lr){2-3} \cmidrule(lr){4-5}
        & Dev          & Test         & Dev           & Test         \\ \cmidrule{2-3} \cmidrule{4-5}
TF-IDF   & 15.4         & 11.2         & 4.8           & 6.5          \\
BOWV    & 14.7         & 18.0         & 4.8           & 4.5          \\
TextCNN & 15.1         & 13.4         & 5.1           & 9.5          \\
BERT    & \textbf{22.9}         & \textbf{21.5}         & \textbf{17.1}          & \textbf{11.6}          \\
Flan-T5  & 19.9            & 14.5            & 13.8             & 5.9            \\
\bottomrule[1.2pt]
\end{tabular}%
}
\caption{Performance comparison on theme keyword identification task across different methods. Metric: macro F1 score.}
\label{tab:theme keyword}
\end{table}

\subsection{Discussion}
All models tested showed low F1 scores in theme identification tasks. This can potentially be attributed to two main factors. Firstly, the limited number of training instances may not provide the models with sufficient information to learn effectively. More importantly, the ambiguous interpretation of theme keywords presents a significant challenge. Achieving consensus, even among human evaluators, can be difficult due to the inherent subjectivity of theme interpretation.

\section{Story-theme matching}
This task is to match one story with the correct theme sentence or vice versa. We formulate it as a text retrieval problem. Take story-theme matching as an example: when given one story, the retrieval model should find the best matching theme sentence out of a collection of all theme sentences. In this setting, a story will act as a query and the themes are the documents. Since there is no ground truth for relevancy scores between every possible story-theme combination, we simply take the original story-theme pair as the only correct match. Therefore, the task is formally described as follows: Given a story (query) $q_i$ and a collection of $N$ theme sentences (documents) $D = \{d_1, d_2, ..., d_i,..., d_N\}$, the learning objective is to rank the correct theme sentence $d_i$ to the top among every other theme sentence in $D$, according to its best relevance to the story (query). Similarly, we switch the story-theme pairs as document-query when using one theme to retrieve the correct story. 

\subsection{Experiments}
In our benchmark analysis, we evaluate the performance of various text retrieval techniques, including conventional and deep methods. Given that we assume a single story has the ground truth theme as its only matching pair and we rank the complete theme sentence set, we employ the Mean Reciprocal Rank (MRR) of the ground truth theme in the retrieval results as our evaluation metric. We use the following calculation, $MRR = \frac{1}{|Q|} \sum_{i=1}^{|Q|} \frac{1}{r_{i}^{d}}$., where $Q$ is the story collection that acts as the queries in the retrieve attempts and $r$ is the ranked position of the gold theme $d$ for one story $q$. 

In this case, a higher ranking $r_{i}$ (represented by smaller numerical values) is considered better. The same setting is applied to theme-story retrieval.

\textbf{BM25:} We use the widely-used BM25 algorithm \cite{jones2000probabilistic} as ranking baseline. 

\textbf{Dense passage retriever (DPR):} We use the architecture of DPR \cite{karpukhin-etal-2020-dense} for the retrieval task: two different BERT encoders are applied for a story (as a query) and a theme sentence (as a document) respectively. A dot-product similarity is used as a ranking metric. The objective is to learn vector representations such that matched story-theme pairs will have higher similarity scores.

\textbf{Sentence-BERT:} Sentence-BERT \cite{reimers-gurevych-2019-sentence} applies to the BERT model in a Siamese network structure. The story and theme sentences are processed independently through the model to get embeddings for similarity calculation.

\textbf{MPnet:} MPnet \cite{song2020mpnet} is trained on both masked and permuted language modeling and shows stronger capability in semantic representations. We apply MPnet in the bi-encoder architecture for this retrieval task.
We train the bi-encoder dense retrievers using contrastive loss with negative samples, i.e., irrelevant theme sentences from other stories. 

\textbf{Cross-Encoder:} We also experiment with a cross-encoder (CE) as a matching scorer. We use MiniLM \cite{wang2020minilm} as our backbone model and perform binary relevance classification between combinations of all the story-theme pairs.

Table \ref{tab:theme match} shows the average rank of the correct theme for a story or the correct story for a given theme in the collection.

\begin{table}[h!]
\centering
\resizebox{0.8\columnwidth}{!}{%
\begin{tabular}{lcccc}
\toprule[1.2pt]
        & \multicolumn{2}{c}{Story-theme} & \multicolumn{2}{c}{Theme-story} \\ \cmidrule(lr){2-3} \cmidrule(lr){4-5}
        & Dev          & Test         & Dev           & Test         \\ \cmidrule{2-3} \cmidrule{4-5}
BM25   & 0.35         & 0.28         & 0.22           & 0.14          \\
DPR    & 0.49        & 0.31         & 0.52           & 0.33          \\
BERT & 0.46         & 0.24         & 0.46           & 0.28          \\
MPnet    & \textbf{0.65}         & \textbf{0.40}         & \textbf{0.60}          & \textbf{0.43}          \\\cmidrule(lr){2-5}
MiniLM ${}_{CE}$   & 0.42	& 0.27         & -         & -    \\
\bottomrule[1.2pt]
\end{tabular}%
}
\caption{Retrieval performance (MRR) of different models on story and theme pairs.}
\label{tab:theme match}
\end{table}

\subsection{Discussion}

In this task, bi-encoders demonstrate substantial efficacy, even under the experimental assumption that each narrative is associated with a single correct theme. The challenges faced by retrieval models appear comparable in both story-theme and theme-story matching. Much like the theme keyword identification task, this task also implies the ambiguity inherent in story interpretation. It is conceivable that a theme (or story) deemed relevant may rank higher than the gold answer.

\section{Story reading comprehension on themes}
Different from many MRC tasks which may contain various contexts and questions, the reading comprehension task on themes is solely targeted at understanding the main idea of the given context. So the question can be uniquely designed like \emph{``What is the main idea of this story?''}, or can be simply omitted. Meanwhile, typical MRC tasks may include span extraction, cloze test, multiple choice, etc., however, story themes are not explicitly reflected in the story context, so problem forms like span extraction and cloze are not applicable. In this work, we design multiple choice problems for reading comprehension on themes: given a story $x$, the model is supposed to identify the correct theme $y$ out of a collection of one correct answer and $N$ distractor themes $\{d_1, d_2, ..., d_N\}$. 

\subsection{Experiments}
To gain a better understanding of the challenges with the reading comprehension of themes in multiple-choice settings, we utilize pretrained Transformer models to address the multiple-choice problems and assess their performance using accuracy as the evaluation metric.

\textbf{BERT and MPnet:} We use the pretrained LMs as in the previous experiments. 
The training of BERT and MPnet follows the same schema: given the story $x$ and a set of options $A = \{a_1, a_2, ..., a_k,..., a_{N+1}\}$, where $a_k$ is either the correct theme sentence $y$ or any distractor $d_i$. The input is formed as the concatenation of the context and one option theme $(x \bigoplus a_i)$. Each of the options will be encoded with the story context and finally produce a set of contextualized representations by taking the \emph{[CLS]} tokens. All of the representations should go through a linear classifier to determine the final answer probability $p({a_1,..., a_{N+1}}|x)$ using the cross-entropy loss.

\textbf{Flan-T5:} For the Text-To-Text multi-task model, Flan-T5, we simply define the input text template as $(x \bigoplus A)$ where every $a_i$ in $A$ is slightly modified by adding a prefix index letter (e.g., A, B, C, ...). Further, the model is trained to generate the correct option letter.

We employ three different strategies for selecting distractors as answer candidates: 
\begin{itemize}
    \item \textbf{Random}: randomly sample distractor candidates from the entire collection of themes;
    \item \textbf{From different virtue class}: select distractor candidates from themes belonging to different virtue categories than the context story; 
    \item \textbf{From same virtue class}: choose distractor candidates from themes that fall within the same virtue category as the context story. 
\end{itemize}
Table \ref{tab:qa} gives the prediction performance comparison of different models under different distractor selection settings.

\begin{table}[h]

\centering
\resizebox{0.9\columnwidth}{!}{%
\begin{tabular}{l|cc|cc|cc}
\toprule[1.2pt]
       & \multicolumn{2}{c|}{Diff Virtue}     & \multicolumn{2}{c|}{Random}   & \multicolumn{2}{c}{Same Virtue}      \\ \cline{2-7} 
       & Dev           & Test          & Dev           & Test          & Dev           & Test          \\ \midrule[1.2pt]
BERT   & \textbf{0.67} & \textbf{0.58} & \textbf{0.74} & \textbf{0.68} & \textbf{0.72} & \textbf{0.64} \\
MPnet  & 0.37          & 0.26          & 0.52          & 0.29          & 0.37          & 0.36          \\
Flan-T5 & 0.56          & 0.46          & 0.50          & 0.41          & 0.56          & 0.46          \\ 
\bottomrule[1.2pt]
\end{tabular}%
}
\caption{Multiple-choice question answering experiment with prediction accuracy score reported.}
\label{tab:qa}
\end{table}

\begin{table*}[h]

\centering
\resizebox{0.80\textwidth}{!}{%
\begin{tabular}{lccc|cccc}
\toprule[1.2pt]

                    & \multicolumn{3}{c|}{Model generated theme} & \multicolumn{4}{c}{Model generated and original theme} \\ \cmidrule(lr){2-4} \cmidrule(lr){5-8}
Source              & Flan-T5   & OPT-175B   & ChatGPT  & Flan-T5   & OPT-175B   & ChatGPT  & Original  \\
Human eval score    & 11                 & 23         & \textbf{50}       & 13                 & 20         & \textbf{46}       & 36        \\
Best interpretation & 7\%                & 17\%       & \textbf{77\%}     & 0\%                & 10\%       & \textbf{50\%}     & 40\%      \\ 
\bottomrule[1.2pt]
\end{tabular}%
}
\caption{Human evaluation of theme generation experiments. We implement two scenarios for human judgment: in the first, only model-generated themes are presented to the human evaluators; in the second, the original theme is included alongside the model-generated ones. A model can achieve a maximum human evaluation score of 60. The best interpretation is expressed as a percentage of the total votes from human judges.}
\label{tab:generation}
\end{table*}

\subsection{Discussion}
In the question-answering task, pretrained LMs demonstrate their ability to leverage learned information from finetuning, with cross-encoding prediction offering the best performance in this context.
We also observe that distractors from the different virtue categories evidently pose a challenge for the best-performing model's predictions, while those from the same categories make the task easier, yet still remain challenging overall.
A common observation from both the matching and multiple-choice Q\&A tasks is that LLMs can identify relevant answers but still fall short of perfect performance. This leaves significant room for further exploration in this area.

\section{Theme generation}
We investigate theme interpretation as a text-generation task. We apply three large-scale LLMs for theme generation on ten held-out stories. The generation adopts a normalized language model prompting schema: A prefix \emph{'Story: '} is added to the story content and a suffix task description and prompt \emph{The main idea of this story is: }.

\textbf{Flan-T5: } We finetune the Text-To-Text language model with story-theme pairs using the template above.

\textbf{OPT-175B: } The OPT-175B model \cite{zhang2022opt} is pretrained on large-scale open-access datasets and has comparable prompting performance to the GPT-3 \cite{brown2020language}.

\textbf{ChatGPT:} OpenAI ChatGPT is an online chatbot service with strong general NLP capabilities. The backbone model is trained with human instructions in a prompt and finetuned with human feedback \cite{ouyang2022training}. 
The version of ChatGPT we experiment with is from February 2023.

Since there is no golden rule to validate the correctness of the generated theme, we use human evaluation on the results. Ten stories are selected as hold-out samples and not used for finetuning. Three human judges are presented with several themes, and asked to perform two tasks after reading a story: 
\begin{enumerate}
    \item Evaluate how reasonable each theme sounds and assign a score. 
    The question for the human judges: \\
    \texttt{Give the rating scores for each main idea based on your judgment. Ratings: 2. Reasonable 1. Somewhat reasonable 0. Not reasonable}. \\
    Given three judges, a maximum of two points that can be earned for each theme, and ten stories in total, the highest score a model can achieve is $3 \times 2 \times 10 = 60$.
    \item Make a single choice on the best-generated theme sentence among the candidates. The question given to the judges is: \\ 
    \texttt{Which answer is the best? <Multiple choice (single answer) question>}.\\
    Each judge gets one vote per story to identify the best theme. We then calculate the percentage of votes each model received relative to the total number of votes to determine its performance.
\end{enumerate}

During the evaluation, each theme's corresponding model name is hidden from the judges, and the order of presentation is randomized. The evaluation results can be found in Table \ref{tab:generation}. We categorize the experiment based on whether the original theme sentence accompanies the generated theme text pieces. Our aim is to determine if human judges might prefer the generated themes, providing a comparative view of the quality of the generated text.

\subsection{Discussion}

In the generation task, we are impressed by the remarkable performance of ChatGPT. The human evaluation shows that it has the capability to produce better quality interpretations than the original theme sentences. The generated interpretations and explanations provide insights into the training strategies used by the model.

A potential reason why human evaluators sometimes favor ChatGPT-generated themes over the original ones could be the inherent simplicity and ambiguity of some original themes. These original themes, often sourced from ancient literature, tend to use concise, philosophical wording that can be challenging to understand for speakers with elementary-level English proficiency. In contrast, ChatGPT articulates in straightforward language and provides a more comprehensive interpretation, which may resonate more with contemporary readers. To ensure a fair comparison, we further refined our experiment, directing ChatGPT to generate a single sentence similar to the original themes. The further designed prompt looks like: \texttt{<Story> Please tell me the main idea of this story. Limit your answer to one single sentence.} Table \ref{tab:generation2} displays the results from this adjusted experimental setup.


\begin{table}[h]

\centering
\resizebox{0.99\columnwidth}{!}{%
\begin{tabular}{l|cccc}
\toprule[1.2pt]

                    & \multicolumn{4}{c}{Model generated and original theme} \\ \cmidrule(lr){2-5}
Source               & Flan-T5   & OPT-175B   & ChatGPT  & Original  \\
Human eval score    & 12                 & 18         & 34       & \textbf{44}        \\
Best interpretation & 3\%                & 13\%       & 30\%     & \textbf{53\%}      \\ 
\bottomrule[1.2pt]
\end{tabular}%
}
\caption{Human judgment on original theme with model-generated themes under one-sentence restriction.}
\label{tab:generation2}
\end{table}

The results indicate that while ChatGPT outperforms other LLM methods, it does not always secure the top preference from human judges. One prominent issue is that when restricted to interpreting the theme in a single sentence, the LLM sometimes defaults to summarizing the story. This shortcoming remains even when the only change in the experimental conditions is the added instruction to provide a one-sentence response. Examples of generated themes and the human-selected best themes are detailed in the Appendix section \ref{appendix:example}. These findings highlight potential limitations in LLMs' capabilities to offer concise theme interpretations. This observation points to a novel research question for future exploration: how can we ensure that LLMs consistently produce quality and reliable theme interpretations within a constrained context window?




\section{Conclusion}
In this work, we first emphasized the importance of advancing NLP models beyond literal comprehension to address more nuanced aspects of reading comprehension, specifically interpretive comprehension. 
This work served as an initial call for the research community to reflect on and further investigate machine reading comprehension (MRC) problems. We introduced theme interpretation as an NLP task in the context of inferential and interpretive reading comprehension, formulating the task comprehensively from various NLP research perspectives and providing extensive empirical research and analysis.
This work builds upon existing MRC research, offering a new perspective by establishing an evaluation criterion for assessing the capability of an NLP system to reason about themes within a narrative content piece.
In the future, we plan to expand the dataset for various research purposes and explore LLM behavior in constrained settings for theme interpretation.

\section*{Limitations}
\textbf{Limitations of the dataset: }
This dataset includes only English versions of stories and corresponding theme texts. The stories from different origins are not balanced in this collection. We acknowledge that many stories originate in Western culture, and contemporary editions might emphasize Western narratives. The sources of the stories are relatively diverse, but they are not balanced and are lacking stories from regions such as Oceania and Africa. Future users of this dataset should be mindful of this limitation in the context of inclusivity and diversity. Despite our best efforts to collect as many story-theme pairs as possible from the open Internet, the literature available for this research topic extends beyond what we've managed to include. We will put effort into further development in future research and invite the community to contribute.

\textbf{Limitations of the usage of LLMs: }
We acknowledge that while current LLMs have limitations, they remain highly effective for executing NLP tasks. Existing research suggests that various prompting and instruction techniques can further enhance the zero-shot capabilities of LLMs for specific tasks. In this work, we choose simple, conversational prompts and instructions, similar to a natural conversation and question-answering, leaving further curated prompting and instruction techniques for future research.

\textbf{Limitations of the annotation process and human evaluation: }
The annotations in the dataset are the result of human effort. We recognize and respect the reality that different people can interpret the same thing in diverse ways. Therefore, our annotations serve as references and are used as experimental ground truths, rather than being considered as universal, absolute truths. We release labels from all annotators and audits with their different opinions for further investigation, not limited to the NLP community.

\section*{Ethics Statement and Broader Impact}
\textbf{Ethical concerns on the dataset: }
We used our best efforts to collect educational stories that are freely accessible on the Internet. A considerable portion of these narratives originate from classical literary works, including \emph{Aesop's Fables} and \emph{Pa\~{n}ca-tantra}. Some of this content may contain values and perspectives that are incompatible with modern sensibilities. Specifically, we discovered that a small number of stories (less than $3\%$) feature explicit, questionable content, such as gender and racial biases as well as discrimination against specific groups of people. Additionally, many fables contain stereotypical portrayals of characters like the 'evil wolf' or 'cunning fox,' which could potentially mislead audiences, particularly children, by oversimplifying the image of specific entities or individuals. We also source stories from contemporary success and entrepreneurial narratives. These narratives might reflect utilitarian values or materialism, and their use for educational purposes should be approached with caution. In our research, we have endeavored to minimize the risk associated with the use of this data by manually labeling and categorizing stories that contain questionable content or potentially raise ethical concerns.


\textbf{Ethical concerns regarding the methodology: }We use large-scale machine learning models (e.g., LLMs) to address the issue of theme interpretation. Present machine learning models carry the risk of inadvertently learning undesired patterns (such as biases) from the narratives in the training data and reproducing these during inference. In potential applications of this research, machine learning models may consequently generate biases originating from the training process. Additionally, the theme sentences provided by authors or editors may not represent diverse perspectives. Given the constraints of the model's training framework, the model may generate main ideas based solely on previously learned resources and knowledge. This could potentially restrict a user's perspective or reinforce the inherent bias in a real-world human-AI interaction scenario.

\textbf{Broader impact: }The broader impact of this work includes theme understanding in other contexts, such as dialogue and video, paving the way for potential research topics and applications, such as automatic lecture interpretation and meeting summarization.

\section*{Acknowledgements}
We thank Tamzid Alam from the University of Houston, Hao Guo from Chalmers University of Technology, and Chaoxian Qi from the University of Houston for their help in data annotation.
We would like to thank the anonymous LREC-COLING reviewers for their feedback on this work. 


\nocite{*}
\section{Bibliographical References}\label{sec:reference}

\bibliographystyle{lrec-coling2024-natbib}
\bibliography{lrec-coling2024-example}

\begin{thebibliography}{0}
\expandafter\ifx\csname natexlab\endcsname\relax\def\natexlab#1{#1}\fi

\end{thebibliography}


\begin{thebibliography}{47}
\expandafter\ifx\csname natexlab\endcsname\relax\def\natexlab#1{#1}\fi

\bibitem[{Aho and Ullman(1972)}]{Aho:72}
Alfred~V. Aho and Jeffrey~D. Ullman. 1972.
\newblock \emph{The Theory of Parsing, Translation and Compiling}, volume~1.
\newblock Prentice-Hall, Englewood Cliffs, NJ.

\bibitem[{{American Psychological Association}(1983)}]{APA:83}
{American Psychological Association}. 1983.
\newblock \emph{Publications Manual}.
\newblock American Psychological Association, Washington, DC.

\bibitem[{Ando and Zhang(2005)}]{Ando2005}
Rie~Kubota Ando and Tong Zhang. 2005.
\newblock \href {https://www.jmlr.org/papers/volume6/ando05a/ando05a.pdf} {A framework for learning predictive structures from multiple tasks and unlabeled data}.
\newblock \emph{Journal of Machine Learning Research}, 6:1817--1853.

\bibitem[{Andrew and Gao(2007)}]{andrew2007scalable}
Galen Andrew and Jianfeng Gao. 2007.
\newblock \href {https://dl.acm.org/doi/abs/10.1145/1273496.1273501} {Scalable training of {$L_1$}-regularized log-linear models}.
\newblock In \emph{Proceedings of the 24th International Conference on Machine Learning}, pages 33--40.

\bibitem[{Brown et~al.(2020)Brown, Mann, Ryder, Subbiah, Kaplan, Dhariwal, Neelakantan, Shyam, Sastry, Askell et~al.}]{brown2020language}
Tom Brown, Benjamin Mann, Nick Ryder, Melanie Subbiah, Jared~D Kaplan, Prafulla Dhariwal, Arvind Neelakantan, Pranav Shyam, Girish Sastry, Amanda Askell, et~al. 2020.
\newblock Language models are few-shot learners.
\newblock \emph{Advances in neural information processing systems}, 33:1877--1901.

\bibitem[{BSI(1973{\natexlab{a}})}]{bs-2570-manual}
BSI. 1973{\natexlab{a}}.
\newblock \emph{Natural Fibre Twines}, 3rd edition.
\newblock British Standards Institution, London.
\newblock BS 2570.

\bibitem[{BSI(1973{\natexlab{b}})}]{bs-2570-techreport}
BSI. 1973{\natexlab{b}}.
\newblock Natural fibre twines.
\newblock BS 2570, British Standards Institution, London.
\newblock 3rd. edn.

\bibitem[{Castor and Pollux(1992)}]{CastorPollux-92}
A.~Castor and L.~E. Pollux. 1992.
\newblock The use of user modelling to guide inference and learning.
\newblock \emph{Applied Intelligence}, 2(1):37--53.

\bibitem[{Chandra et~al.(1981)Chandra, Kozen, and Stockmeyer}]{Chandra:81}
Ashok~K. Chandra, Dexter~C. Kozen, and Larry~J. Stockmeyer. 1981.
\newblock \href {https://doi.org/10.1145/322234.322243} {Alternation}.
\newblock \emph{Journal of the Association for Computing Machinery}, 28(1):114--133.

\bibitem[{Chercheur(1994)}]{Chercheur-94}
J.L. Chercheur. 1994.
\newblock \emph{Case-Based Reasoning}, 2nd edition.
\newblock Morgan Kaufman Publishers, San Mateo, CA.

\bibitem[{Chomsky(1973)}]{chomsky-73}
N.~Chomsky. 1973.
\newblock Conditions on transformations.
\newblock In \emph{A festschrift for {Morris Halle}}, New York. Holt, Rinehart \& Winston.

\bibitem[{Chung et~al.(2022)Chung, Hou, Longpre, Zoph, Tay, Fedus, Li, Wang, Dehghani, Brahma et~al.}]{chung2022scaling}
Hyung~Won Chung, Le~Hou, Shayne Longpre, Barret Zoph, Yi~Tay, William Fedus, Eric Li, Xuezhi Wang, Mostafa Dehghani, Siddhartha Brahma, et~al. 2022.
\newblock Scaling instruction-finetuned language models.
\newblock \emph{arXiv preprint arXiv:2210.11416}.

\bibitem[{Cooley and Tukey(1965)}]{ct1965}
James~W. Cooley and John~W. Tukey. 1965.
\newblock \href {https://www.ams.org/journals/mcom/1965-19-090/S0025-5718-1965-0178586-1/S0025-5718-1965-0178586-1.pdf} {An algorithm for the machine calculation of complex {F}ourier series}.
\newblock \emph{Mathematics of Computation}, 19(90):297--301.

\bibitem[{Devlin et~al.(2019)Devlin, Chang, Lee, and Toutanova}]{devlin-etal-2019-bert}
Jacob Devlin, Ming-Wei Chang, Kenton Lee, and Kristina Toutanova. 2019.
\newblock \href {https://doi.org/10.18653/v1/N19-1423} {{BERT}: Pre-training of deep bidirectional transformers for language understanding}.
\newblock In \emph{Proceedings of the 2019 Conference of the North {A}merican Chapter of the Association for Computational Linguistics: Human Language Technologies, Volume 1 (Long and Short Papers)}, pages 4171--4186, Minneapolis, Minnesota. Association for Computational Linguistics.

\bibitem[{Eco(1990)}]{Eco:1990}
Umberto Eco. 1990.
\newblock \emph{The Limits of Interpretation}.
\newblock Indian University Press.

\bibitem[{Gusfield(1997)}]{Gusfield:97}
Dan Gusfield. 1997.
\newblock \href {https://www.cambridge.org/core/books/algorithms-on-strings-trees-and-sequences/F0B095049C7E6EF5356F0A26686C20D3} {\emph{Algorithms on Strings, Trees and Sequences}}.
\newblock Cambridge University Press, Cambridge, UK.

\bibitem[{Herber(1978)}]{herber1978}
Harold~L. Herber. 1978.
\newblock \emph{Teaching reading in content areas}, 2 edition.
\newblock Prentice-Hall, Englewood Cliffs, N.J.

\bibitem[{Hermans et~al.(2017)Hermans, Beyer, and Leibe}]{hermans2017defense}
Alexander Hermans, Lucas Beyer, and Bastian Leibe. 2017.
\newblock In defense of the triplet loss for person re-identification.
\newblock \emph{arXiv preprint arXiv:1703.07737}.

\bibitem[{Hoel(1971{\natexlab{a}})}]{hoel-71-whole}
Paul~Gerhard Hoel. 1971{\natexlab{a}}.
\newblock \emph{Elementary Statistics}, 3rd edition.
\newblock Wiley series in probability and mathematical statistics. Wiley, New York, Chichester.
\newblock ISBN 0~471~40300.

\bibitem[{Hoel(1971{\natexlab{b}})}]{hoel-71-portion}
Paul~Gerhard Hoel. 1971{\natexlab{b}}.
\newblock \emph{Elementary Statistics}, 3rd edition, Wiley series in probability and mathematical statistics, pages 19--33. Wiley, New York, Chichester.
\newblock ISBN 0~471~40300.

\bibitem[{Hu et~al.(2021)Hu, Shen, Wallis, Allen-Zhu, Li, Wang, Wang, and Chen}]{hu2021lora}
Edward~J Hu, Yelong Shen, Phillip Wallis, Zeyuan Allen-Zhu, Yuanzhi Li, Shean Wang, Lu~Wang, and Weizhu Chen. 2021.
\newblock Lora: Low-rank adaptation of large language models.
\newblock \emph{arXiv preprint arXiv:2106.09685}.

\bibitem[{Jespersen(1922)}]{Jespersen:1922}
Otto Jespersen. 1922.
\newblock \emph{Language: Its Nature, Development, and Origin}.
\newblock Allen and Unwin.

\bibitem[{Jones et~al.(2000)Jones, Walker, Robertson, and Robertson}]{jones2000probabilistic}
K~Sparck Jones, S~Walker, SE~Robertson, and Stephen Robertson. 2000.
\newblock A probabilistic model of information retrieval: development and comparative experiments.
\newblock \emph{Information Processing and Management}, 36:779--808.

\bibitem[{Karpukhin et~al.(2020)Karpukhin, Oguz, Min, Lewis, Wu, Edunov, Chen, and Yih}]{karpukhin-etal-2020-dense}
Vladimir Karpukhin, Barlas Oguz, Sewon Min, Patrick Lewis, Ledell Wu, Sergey Edunov, Danqi Chen, and Wen-tau Yih. 2020.
\newblock \href {https://doi.org/10.18653/v1/2020.emnlp-main.550} {Dense passage retrieval for open-domain question answering}.
\newblock In \emph{Proceedings of the 2020 Conference on Empirical Methods in Natural Language Processing (EMNLP)}, pages 6769--6781, Online. Association for Computational Linguistics.

\bibitem[{Kim(2014)}]{kim-2014-convolutional}
Yoon Kim. 2014.
\newblock \href {https://doi.org/10.3115/v1/D14-1181} {Convolutional neural networks for sentence classification}.
\newblock In \emph{Proceedings of the 2014 Conference on Empirical Methods in Natural Language Processing ({EMNLP})}, pages 1746--1751, Doha, Qatar. Association for Computational Linguistics.

\bibitem[{Ko{\v{c}}isk{\'y} et~al.(2018)Ko{\v{c}}isk{\'y}, Schwarz, Blunsom, Dyer, Hermann, Melis, and Grefenstette}]{kocisky-etal-2018-narrativeqa}
Tom{\'a}{\v{s}} Ko{\v{c}}isk{\'y}, Jonathan Schwarz, Phil Blunsom, Chris Dyer, Karl~Moritz Hermann, G{\'a}bor Melis, and Edward Grefenstette. 2018.
\newblock \href {https://doi.org/10.1162/tacl_a_00023} {The {N}arrative{QA} reading comprehension challenge}.
\newblock \emph{Transactions of the Association for Computational Linguistics}, 6:317--328.

\bibitem[{Mangrulkar et~al.(2022)Mangrulkar, Gugger, Debut, Belkada, Paul, and Bossan}]{peft}
Sourab Mangrulkar, Sylvain Gugger, Lysandre Debut, Younes Belkada, Sayak Paul, and Benjamin Bossan. 2022.
\newblock Peft: State-of-the-art parameter-efficient fine-tuning methods.
\newblock \url{https://github.com/huggingface/peft}.

\bibitem[{Nguyen et~al.(2016)Nguyen, Rosenberg, Song, Gao, Tiwary, Majumder, and Deng}]{nguyen2016ms}
Tri Nguyen, Mir Rosenberg, Xia Song, Jianfeng Gao, Saurabh Tiwary, Rangan Majumder, and Li~Deng. 2016.
\newblock Ms marco: A human generated machine reading comprehension dataset.
\newblock \emph{choice}, 2640:660.

\bibitem[{Ouyang et~al.(2022)Ouyang, Wu, Jiang, Almeida, Wainwright, Mishkin, Zhang, Agarwal, Slama, Ray et~al.}]{ouyang2022training}
Long Ouyang, Jeffrey Wu, Xu~Jiang, Diogo Almeida, Carroll Wainwright, Pamela Mishkin, Chong Zhang, Sandhini Agarwal, Katarina Slama, Alex Ray, et~al. 2022.
\newblock Training language models to follow instructions with human feedback.
\newblock \emph{Advances in Neural Information Processing Systems}, 35:27730--27744.

\bibitem[{Pennington et~al.(2014)Pennington, Socher, and Manning}]{pennington-etal-2014-glove}
Jeffrey Pennington, Richard Socher, and Christopher Manning. 2014.
\newblock \href {https://doi.org/10.3115/v1/D14-1162} {{G}lo{V}e: Global vectors for word representation}.
\newblock In \emph{Proceedings of the 2014 Conference on Empirical Methods in Natural Language Processing ({EMNLP})}, pages 1532--1543, Doha, Qatar. Association for Computational Linguistics.

\bibitem[{Peterson et~al.(2004)Peterson, Seligman et~al.}]{peterson2004character}
Christopher Peterson, Martin~EP Seligman, et~al. 2004.
\newblock \emph{Character strengths and virtues: A handbook and classification}, volume~1.
\newblock Oxford University Press.

\bibitem[{Rajpurkar et~al.(2016)Rajpurkar, Zhang, Lopyrev, and Liang}]{rajpurkar-etal-2016-squad}
Pranav Rajpurkar, Jian Zhang, Konstantin Lopyrev, and Percy Liang. 2016.
\newblock \href {https://doi.org/10.18653/v1/D16-1264} {{SQ}u{AD}: 100,000+ questions for machine comprehension of text}.
\newblock In \emph{Proceedings of the 2016 Conference on Empirical Methods in Natural Language Processing}, pages 2383--2392, Austin, Texas. Association for Computational Linguistics.

\bibitem[{Rasooli and Tetreault(2015)}]{rasooli-tetrault-2015}
Mohammad~Sadegh Rasooli and Joel~R. Tetreault. 2015.
\newblock \href {http://arxiv.org/abs/1503.06733} {Yara parser: {A} fast and accurate dependency parser}.
\newblock \emph{Computing Research Repository}, arXiv:1503.06733.
\newblock Version 2.

\bibitem[{Reimers and Gurevych(2019)}]{reimers-gurevych-2019-sentence}
Nils Reimers and Iryna Gurevych. 2019.
\newblock \href {https://doi.org/10.18653/v1/D19-1410} {Sentence-{BERT}: Sentence embeddings using {S}iamese {BERT}-networks}.
\newblock In \emph{Proceedings of the 2019 Conference on Empirical Methods in Natural Language Processing and the 9th International Joint Conference on Natural Language Processing (EMNLP-IJCNLP)}, pages 3982--3992, Hong Kong, China. Association for Computational Linguistics.

\bibitem[{Richardson et~al.(2013)Richardson, Burges, and Renshaw}]{richardson-etal-2013-mctest}
Matthew Richardson, Christopher~J.C. Burges, and Erin Renshaw. 2013.
\newblock \href {https://aclanthology.org/D13-1020} {{MCT}est: A challenge dataset for the open-domain machine comprehension of text}.
\newblock In \emph{Proceedings of the 2013 Conference on Empirical Methods in Natural Language Processing}, pages 193--203, Seattle, Washington, USA. Association for Computational Linguistics.

\bibitem[{Saha et~al.(2018)Saha, Aralikatte, Khapra, and Sankaranarayanan}]{saha-etal-2018-duorc}
Amrita Saha, Rahul Aralikatte, Mitesh~M. Khapra, and Karthik Sankaranarayanan. 2018.
\newblock \href {https://doi.org/10.18653/v1/P18-1156} {{D}uo{RC}: Towards complex language understanding with paraphrased reading comprehension}.
\newblock In \emph{Proceedings of the 56th Annual Meeting of the Association for Computational Linguistics (Volume 1: Long Papers)}, pages 1683--1693, Melbourne, Australia. Association for Computational Linguistics.

\bibitem[{Singer et~al.(1954--58)Singer, Holmyard, and Hall}]{singer-whole}
Charles~Joseph Singer, E.~J. Holmyard, and A.~R. Hall, editors. 1954--58.
\newblock \emph{A history of technology}.
\newblock Oxford University Press, London.
\newblock 5 vol.

\bibitem[{Song et~al.(2020)Song, Tan, Qin, Lu, and Liu}]{song2020mpnet}
Kaitao Song, Xu~Tan, Tao Qin, Jianfeng Lu, and Tie-Yan Liu. 2020.
\newblock Mpnet: Masked and permuted pre-training for language understanding.
\newblock \emph{Advances in Neural Information Processing Systems}, 33:16857--16867.

\bibitem[{Strötgen and Gertz(2012)}]{Martin-90}
Jannik Strötgen and Michael Gertz. 2012.
\newblock Temporal tagging on different domains: Challenges, strategies, and gold standards.
\newblock In \emph{Proceedings of the Eight International Conference on Language Resources and Evaluation (LREC'12)}, pages 3746--3753, Istanbul, Turkey. European Language Resource Association (ELRA).

\bibitem[{Superman et~al.(2000)Superman, Batman, Catwoman, and Spiderman}]{Superman-Batman-Catwoman-Spiderman-00}
S.~Superman, B.~Batman, C.~Catwoman, and S.~Spiderman. 2000.
\newblock \emph{Superheroes experiences with books}, 20th edition.
\newblock The Phantom Editors Associates, Gotham City.

\bibitem[{Vacca and Vacca(1998)}]{vacca1998}
Richard~T Vacca and Jo~Anne~L Vacca. 1998.
\newblock \emph{Content area reading: {L}iteracy and learning across the curriculum}, 6 edition.
\newblock Longman.

\bibitem[{Wang et~al.(2020)Wang, Wei, Dong, Bao, Yang, and Zhou}]{wang2020minilm}
Wenhui Wang, Furu Wei, Li~Dong, Hangbo Bao, Nan Yang, and Ming Zhou. 2020.
\newblock Minilm: Deep self-attention distillation for task-agnostic compression of pre-trained transformers.
\newblock \emph{Advances in Neural Information Processing Systems}, 33:5776--5788.

\bibitem[{Xu et~al.(2022)Xu, Wang, Yu, Ritchie, Yao, Wu, Zhang, Li, Bradford, Sun, Hoang, Sang, Hou, Ma, Yang, Peng, Yu, and Warschauer}]{xu-etal-2022-fantastic}
Ying Xu, Dakuo Wang, Mo~Yu, Daniel Ritchie, Bingsheng Yao, Tongshuang Wu, Zheng Zhang, Toby Li, Nora Bradford, Branda Sun, Tran Hoang, Yisi Sang, Yufang Hou, Xiaojuan Ma, Diyi Yang, Nanyun Peng, Zhou Yu, and Mark Warschauer. 2022.
\newblock \href {https://doi.org/10.18653/v1/2022.acl-long.34} {Fantastic questions and where to find them: {F}airytale{QA} {--} an authentic dataset for narrative comprehension}.
\newblock In \emph{Proceedings of the 60th Annual Meeting of the Association for Computational Linguistics (Volume 1: Long Papers)}, pages 447--460, Dublin, Ireland. Association for Computational Linguistics.

\bibitem[{Yang et~al.(2024)Yang, Zhou, Wong, and Zhang}]{yang2024loretta}
Yifan Yang, Jiajun Zhou, Ngai Wong, and Zheng Zhang. 2024.
\newblock Loretta: Low-rank economic tensor-train adaptation for ultra-low-parameter fine-tuning of large language models.
\newblock \emph{arXiv preprint arXiv:2402.11417}.

\bibitem[{Zhang et~al.(2022)Zhang, Roller, Goyal, Artetxe, Chen, Chen, Dewan, Diab, Li, Lin et~al.}]{zhang2022opt}
Susan Zhang, Stephen Roller, Naman Goyal, Mikel Artetxe, Moya Chen, Shuohui Chen, Christopher Dewan, Mona Diab, Xian Li, Xi~Victoria Lin, et~al. 2022.
\newblock Opt: Open pre-trained transformer language models.
\newblock \emph{arXiv preprint arXiv:2205.01068}.

\bibitem[{Zhang et~al.(2021)Zhang, Shafaei, Gonzalez, and Solorio}]{zhang-etal-2021-none-severe}
Yigeng Zhang, Mahsa Shafaei, Fabio Gonzalez, and Thamar Solorio. 2021.
\newblock \href {https://doi.org/10.18653/v1/2021.findings-emnlp.332} {From none to severe: {P}redicting severity in movie scripts}.
\newblock In \emph{Findings of the Association for Computational Linguistics: EMNLP 2021}, pages 3951--3956, Punta Cana, Dominican Republic. Association for Computational Linguistics.

\bibitem[{Zhao et~al.(2023)Zhao, Kim, Liu, Piramuthu, and Hakkani-Tür}]{Zhao2023storyqa}
Sanqiang Zhao, Seokhwan Kim, Yang Liu, Robinson Piramuthu, and Dilek Hakkani-Tür. 2023.
\newblock \href {https://www.amazon.science/publications/storyqa-story-grounded-question-answering-dataset} {Storyqa: Story grounded question answering dataset}.
\newblock In \emph{AAAI 2023 Workshop on Knowledge Augmented Methods for NLP}.

\end{thebibliography}

\label{lr:ref}
\bibliographystylelanguageresource{lrec-coling2024-natbib}
\bibliographylanguageresource{languageresource}

\section{Appendix}
\label{sec:appendix}
\subsection{Details about the human annotators, auditors, and judges}\label{appendix:annotation}
The individuals involved in the theme keyword identification task, as well as the evaluation of the generated text, were all adults with advanced proficiency in English (qualified for graduate school study in English-speaking countries). They demonstrated a comprehensive understanding of the instructions, as well as the story content necessary for annotation, auditing, and evaluation processes. We have obtained consent from these individuals to use the data and results they provided in this paper.
\subsection{Implementation and experiment details} \label{appendix:experiment}
This subsection outlines the details of the methods utilized in the different tasks and the settings of the experiments. For the tasks of theme keyword identification, story-theme matching, and story reading comprehension on themes, there are $20\%$ of instances are held out for testing, and $15\%$ of training instances are used for validation. Recent advances have demonstrated the effectiveness of parameter-efficient methods for language model tuning, as seen in works like \cite{hu2021lora,peft,yang2024loretta}. However, the primary focus of our work is on evaluating performance on specific tasks, and given that the model size is still in reasonable size for finetuning, we choose to finetune using the full set of parameters for all models to be trained in this work. More detailed information can be found in our project repository.
\subsubsection{Theme keyword identification}
In this task, we performed two classification experiments related to character virtues and strengths. There are six categories for virtues, and 14 out of 24 categories for strengths have no fewer than three instances each split for training, validation, and testing. 

The GloVe embedding vectors used in this task were trained on 840B tokens from Common Crawl, and consist of 300 dimensions. The TextCNN model uses the same GloVe embedding, and its CNN modules have kernel sizes of 3, 4, and 5. The BERT-base and Flan-T5 base models for fine-tuning in this task are adapted from HuggingFace. The configuration settings for the TextCNN and BERT baseline models were mostly adapted from an existing study that focused on analyzing story flow data through movie dialogue scripts \cite{zhang-etal-2021-none-severe}.

We also report the classification performance of a proprietary model, \texttt{GPT-3.5-turbo}, since the ChatGPT version from February 2023 (used in the theme generation task) is no longer trackable. The zero-shot prompting performance on the test set from an experimental run on this task is shown in Table \ref{tab:theme keyword chatgpt}.
\begin{table}[h]

\centering
\resizebox{0.6\columnwidth}{!}{%
\begin{tabular}{l|cc}
\toprule[1.2pt]

Granularity           & Virtue  & Strength  \\ 
F1 score        & 28.7       & 17.0     \\
\bottomrule[1.2pt]
\end{tabular}%
}
\caption{Zero-shot classification performance using \texttt{GPT-3.5-turbo}.}
\label{tab:theme keyword chatgpt}
\end{table}

\subsubsection{Story-theme matching}
In this task, we employ DPR, Sentence-BERT, and MPNet in a bi-encoder setting for encoding the story and theme respectively (or vice versa). These backbone models are trained on different tasks such as question-answer retrieval and search document retrieval to acquire capabilities in semantic textual similarity. The fine-tuning process involves multiple negative samples to learn strong representations \cite{hermans2017defense}. The cross-encoder model in this task is a MiniLM model trained on the MS Marco Passage Ranking task \cite{nguyen2016ms}.
\subsubsection{Story reading comprehension on themes }
To ensure a fair comparison by maintaining roughly the same model size, we use the BERT-base, MPNet-base, and Flan-T5-small models for multiple-choice task fine-tuning. We manually augment different multiple-choice candidates to each story five times during training to obtain better learning outcomes.

Here we also report results on the test set from an experimental run with \texttt{GPT-3.5-turbo} in a zero-shot manner in Table \ref{tab:qa chatgpt}.
\begin{table}[h]

\centering
\resizebox{0.9\columnwidth}{!}{%
\begin{tabular}{l|ccc}
\toprule[1.2pt]

Difficulty           & Diff Virtue  & Random & Same Virtue \\ 
Accuracy        & 0.78       & 0.76   & 0.69  \\
\bottomrule[1.2pt]
\end{tabular}%
}
\caption{Zero-shot multiple-choice question answering performance using \texttt{GPT-3.5-turbo}.}
\label{tab:qa chatgpt}
\end{table}

\subsubsection{Theme generation}
We select ten stories from different sources as a held-out test set and use the rest for model fine-tuning. We use the Flan-T5 large model ($\sim 0.8B$) to perform instruction tuning with the theme generation objective on the training set for three epochs. The OPT-175B model generation result was collected from the Colossal-AI web service. The ChatGPT results date from February 2023.

\subsection{Examples of theme generation and evaluation}\label{appendix:example}
We list three examples of story theme generation outcomes and their human evaluation results from the version of evaluation that \emph{includes} the original theme. ChatGPT is enforced to generate one single theme sentence. The symbol \textbf{$\surd$} indicates the one that was chosen to be the best theme sentence by the human judges.
\subsubsection{Story 1}
A Fox fell into a well, and though it was not very deep, he found that he could not get out again. After he had been in the well a long time, a thirsty Goat came by. The Goat thought the Fox had gone down to drink, and so he asked if the water was good. "The finest in the whole country," said the crafty Fox, "jump in and try it. There is more than enough for both of us." The thirsty Goat immediately jumped in and began to drink. The Fox just as quickly jumped on the Goat's back and leaped from the tip of the Goat's horns out of the well The foolish Goat now saw what a plight he had got into, and begged the Fox to help him out. But the Fox was already on his way to the woods "If you had as much sense as you have beard, old fellow," he said as he ran, "you would have been more cautious about finding a way to get out again before you jumped in."

The main idea of the story is: 

\begin{enumerate}
    \item \textbf{Flan-T5}: The wisest people are the ones who are not fools.
    \item \textbf{OPT-175B}: A man should not be too quick to take advantage of an opportunity.[irrelevant trailing content]
    \item \textbf{ChatGPT}: The clever Fox outwits the foolish Goat and uses him to escape the well.
    \item \textbf{Original}: Look before you leap. \textbf{$\surd$}
\end{enumerate}

\subsubsection{Story 2}
One bright day in late autumn a family of Ants were bustling about in the warm sunshine, drying out the grain they had stored up during the summer, when a starving Grasshopper, his fiddle under his arm, came up and humbly begged for a bite to eat. "What!" cried the Ants in surprise, "haven't you stored anything away for the winter? What in the world were you doing all last summer?" "I didn't have time to store up any food," whined the Grasshopper; "I was so busy making music that before I knew it the summer was gone." The Ants shrugged their shoulders in disgust. "Making music, were you?" they cried. "Very well; now dance!" And they turned their backs on the Grasshopper and went on with their work.

The main idea of the story is:

\begin{enumerate}
    \item \textbf{Flan-T5}: The Grasshopper is not a good mate.
    \item \textbf{OPT-175B}: Every one of us has different capacities and inclinations.  The ants were busy storing up food for the winter, but the grasshopper did not. 
    \item \textbf{ChatGPT}: Preparation is key and those who fail to plan ahead will face consequences. \textbf{$\surd$}
    \item \textbf{Original}: There's a time for work and a time for play.
\end{enumerate}

\subsubsection{Story 3}
Once a rich merchant’s house was robbed. The merchant suspected that the thief was one of his servants. So he went to Birbal and mentioned the incident. Birbal went to his house and assembled all of his servants and asked that who stole the merchant’s things. Everybody denied. Birbal thought for a moment, then gave a stick of equal length to all the servants of the merchant and said to them that the stick of the real thief will be longer by two inches tomorrow. All the servants should be present here again tomorrow with their sticks. All the servants went to their homes and gathered again at the same place the next day. Birbal asked them to show him their sticks. One of the servants had his stick shorter by two inches. Birbal said, “This is your thief, merchant.” Later the merchant asked Birbal, “How did you catch him?” Birbal said, “The thief had already cut his stick short by two inches in the night fearing that his stick will be longer by two inches by morning.”

The main idea of the story is: 
\begin{enumerate}
    \item \textbf{Flan-T5}: The thief is always a thief.
    \item \textbf{OPT-175B}: A. If you want to find out the thief, first punish all the servants; B. The real thief is not afraid of punishment;
    \item \textbf{ChatGPT}: Birbal solving a robbery case using clever reasoning.
    \item \textbf{Original}: Truth will always Prevail. \textbf{$\surd$}
\end{enumerate}

\end{document}